%% file: acl_latex.tex
\pdfoutput=1

\documentclass[11pt]{article}
\usepackage[utf8]{inputenc}
\usepackage{multirow}
\usepackage{graphicx}

\usepackage[table]{xcolor} 
\usepackage[final]{acl}

\usepackage{comment} 

\usepackage{booktabs}

\usepackage{kotex}

\usepackage{times}

\usepackage{latexsym}

\usepackage[T1]{fontenc}

\usepackage[utf8]{inputenc}

\usepackage{microtype}

\usepackage{inconsolata}

\usepackage{graphicx}

\title{Team ACK at SemEval-2025 Task 2:

Beyond Word-for-Word Machine Translation for English-Korean Pairs}

\author{Daniel Lee\thanks{\ These authors contributed equally.} \\
	Adobe Inc.\\
	\texttt{dlee1@adobe.com} \\\And{}
	Harsh Sharma\footnotemark[1] \\
	CU Boulder \\
	\texttt{harsh.sharma@colorado.edu} \\\And{}
	Jieun Han \\
	KAIST \\
	\texttt{jieun\_han@kaist.ac.kr} \\\AND{}
	Sunny Jeong \\
	New York University \\
	\texttt{sunny.jeong@nyu.edu} \\\And{}
        Alice Oh \\
        KAIST \\ 
        \texttt{alice.oh@kaist.edu} \\\And{}
        Vered Shwartz \\
        UBC \\ 
        \texttt{vshwartz@cs.ubc.ca} \\
}

\begin{document}
\maketitle
\begin{abstract}
Translating knowledge-intensive and entity-rich text between English and Korean requires transcreation to preserve language-specific and cultural nuances beyond literal, phonetic or word-for-word conversion. We evaluate 13 models (LLMs and MT models) using automatic metrics and human assessment by bilingual annotators. Our findings show LLMs outperform traditional MT systems but struggle with entity translation requiring cultural adaptation. By constructing an error taxonomy, we identify incorrect responses and entity name errors as key issues, with performance varying by entity type and popularity level. This work exposes gaps in automatic evaluation metrics and hope to enable future work in completing culturally-nuanced machine translation.
\end{abstract}

\section{Introduction}
Machine Translation (MT) has progressed significantly with the introduction of the transformer paradigm \cite{wang-etal-2023-document-level}. Supervised machine translation models have improved their performance in challenging scenarios such as long-document translation and stylized translation \cite{lyu-etal-2024-paradigm}. Despite the success of these models in general translation, they still struggle to translate named entities which are culturally-naunced or language-specific~\cite{10.1007/s10994-021-06073-9}, in other words, entities which are rooted in social, geographic, historical, and political contexts~\cite{hershcovich-etal-2022-challenges}. However, the emergence of self-supervised large language models (LLMs) and their zero-shot translation capabilities have shown to be a promising avenue to address these problems. Their ability to learn in-context enables new capabilities, such as terminology constrained translation unlike foundation approaches~\cite{koshkin-etal-2024-llms}.

In this paper, we conduct a thorough analysis of English-to-Korean translation through the following:
\begin{itemize}
    \item We conduct a comprehensive evaluation of 13 models (including LLMs and traditional MT models) on English-Korean translation pairs, focusing specifically on knowledge-intensive and entity-dense text.
    \item We complete a thorough human evaluation with bilingual annotators to construct a comprehensive error taxonomy.
    \item We reveal important gaps in automatic evaluation metrics (comparing BLEU, COMET, and M-ETA scores against human assessments), demonstrating that these metrics often fail to capture cultural and linguistic nuances in entity translation.
\end{itemize}
We hope this focused work on English-Korean motivates similar work in other domains, to further understand culturally-nuanced and language-specific translation.

\section{Related Work}

While MT has continued to improve from RNN-based models \cite{NIPS20145a18e133} to transformer-based models \citep[][\textit{inter alia}]{koishekenov-etal-2023-memory, tang2020multilingualtranslationextensiblemultilingual, zhu-etal-2024-multilingual-contrastive, alves-etal-2023-steering, wang-etal-2023-document-level, zaranis-etal-2024-analyzing}, entity translation remains a significant challenge due to the need for both direct word-for-word conversion (\emph{transliteration}) and contextual adaptation (\emph{transcreation}) \cite{hershcovich-etal-2022-challenges}. For example, the English query, "\textit{What is the Rotten Tomatoes score of John Wick?}" should result in "\textit{Rotten Tomatoes}" being translated to "\textit{로튼 토마토}", the movies and TV review site, instead of "\textit{썩은 토마토}", the literal translation meaning rotting fruit.
While LLMs are a promising avenue to address these problems, they are primarily trained on large-scale multilingual corpora with an English-centric bias. This results in them struggling to capture the nuanced sociocultural and historical contexts necessary for effective transcreation \cite{ponti-etal-2020-xcopa}.
Efforts to enhance entity translation through retrieval-augmented generation (RAG) have been introduced, such as leveraging knowledge graphs (KGs) and structured databases. For example, KG-MT proposed using multilingual knowledge graphs to improve cultural adaptation in MT by providing contextually appropriate entity translations \cite{conia-etal-2024-towards}. 

The complexities of English-Korean entity translation stem from fundamental linguistic and cultural differences between the two languages. 

Transliteration is complicated by phonetic variations and word structure differences, while transcreation requires adapting names, idioms, and references to maintain cultural and linguistic naturalness \cite{pedersen2014exploring, diaz2023towards}. 
Existing research has primarily focused on Western-centric language pairs, leaving English-Korean entity translation an area in need of further investigation \cite{kim-choi-2015-entity,kim-etal-2022-kochet}.
Given these challenges, improving LLM-driven MT requires context-sensitive modeling and culturally aware translation strategies. 
This study aims to bridge these gaps by evaluating state-of-the-art LLMs and multilingual MT models on entity-dense and knowledge-intensive texts, combining automatic evaluation metrics with human assessment to gain insights into translation quality.

\section{Experimental Setup}
To comprehensively evaluate the performance of LLMs on the task of MT, we consider 13 models including the most popular and best performing LLMs from OpenAI (GPT-4, GPT-4o, o1, o1-mini), Anthropic (Claude 3.5 Sonnet, 3.5 Haiku), Google (Gemini 1.5 Flash, 1.5 Pro), Meta (Llama3-8B), Grok (grok-2), and DeepSeek (R1-7B) and recent multilingual MT models (NLLB-200 and mBART-50) \cite{openai_o1_system_card, openai_gpt4o_system_card, anthropic_claude35_computer_use, gemini15_report, xai_grok_1212, deepseekai2025deepseekr1incentivizingreasoningcapability, grattafiori2024llama3herdmodels, koishekenov-etal-2023-memory, tang2020multilingualtranslationextensiblemultilingual}. With these models, we conduct an automatic evaluation with several metrics (Section~\ref{autoeval}) and an in-depth human evaluation (Section~\ref{humaneval}). 
For this evaluation, we use the task dataset provided by \newcite{conia-etal-2025-semeval-2025} that was prepared from XC-Translate \cite{conia-etal-2024-towards}. XC-Translate is a multi-reference, human-curated dataset that is challenging due to its focus on translating cross-cultural texts containing entity names.

\input{tables/automaticeval}

\subsection{Automatic Evaluation}
\label{autoeval}
The automatic evaluation was conducted on the same 5,082 English-Korean pairs for each model, which comprised of several machine translation metrics including:

\begin{itemize}
    \item \textbf{BLEU:} Measures the n-gram overlap between the translated text against the reference translations  \cite{10.3115/1073083.1073135}.
    \item \textbf{COMET:} Predicts human judgments of machine translation quality using neural models \cite{rei-etal-2020-comet}.
    \item \textbf{M-ETA:} Measures the translation quality at the entity level \cite{conia-etal-2024-towards}.
\end{itemize}

\begin{figure*}[t]
        \centering
        \includegraphics[width=.8\textwidth]{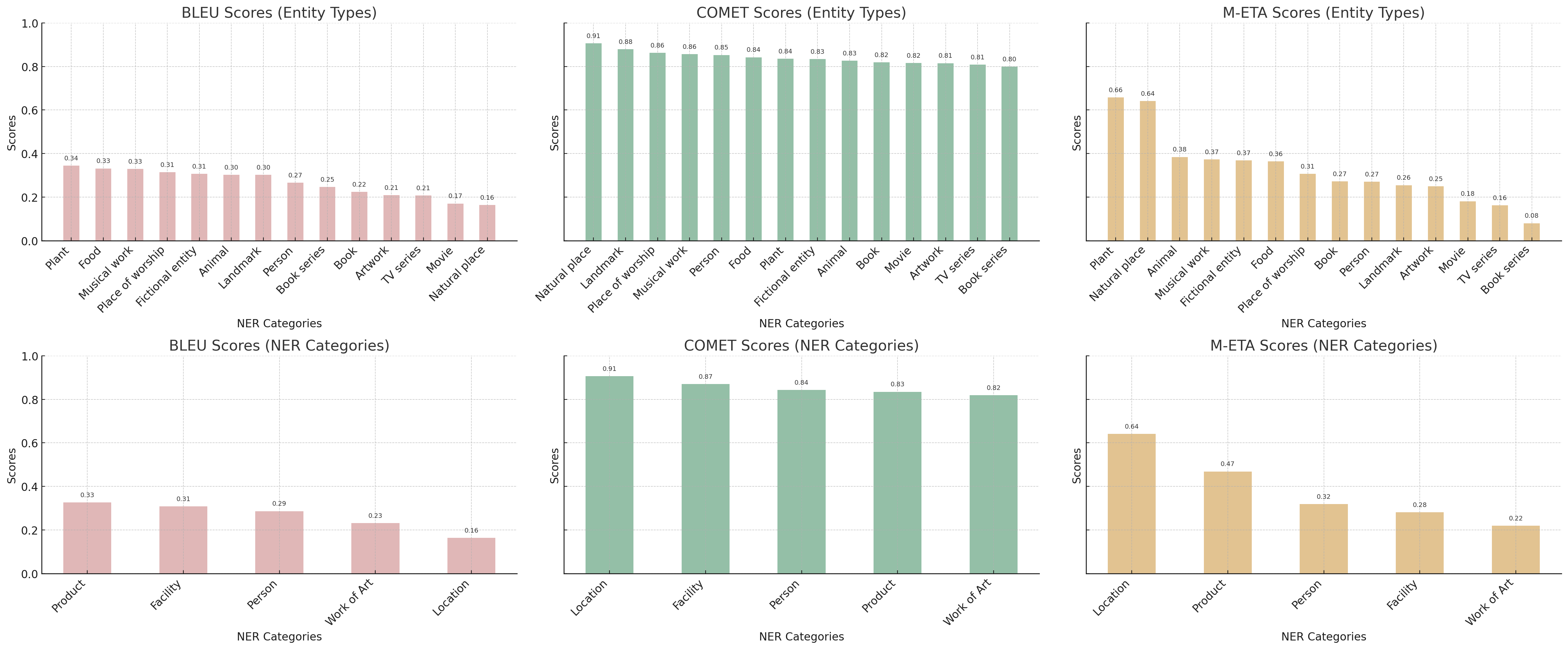}
    \caption{Average BLEU, COMET, M-ETA scores by entity types.}
    \label{fig:ner_entity_average}
\end{figure*}

We use a combination of the three automatic metrics augmented by human evaluation as they individually do not capture the nuances of machine translation. BLEU, while the industry standard and resource efficient, lacks strong correlation to human judgment \cite{callison-burch-etal-2006-evaluating}. COMET better correlates with human judgments thanks to moving beyond n-grams to semantic understanding, yet it does not reveal fine-grained word-level insights \cite{kaffee_et_al:TGDK.1.1.10} like culturally or language appropriate entities. M-ETA adds to both by focusing (only) on entity-level translation quality. Finally, human evaluation can detect translation errors not captured by the automatic metrics and provide in-depth feedback. Due to its cost-intensive and time-consuming nature, we only manually evaluate a portion of the data.  
Used together, these metrics address the limitations proposed by each approach for a comprehensive analysis of machine translation quality.
\begin{figure*}[t]
        \centering
        \includegraphics[width=\textwidth]{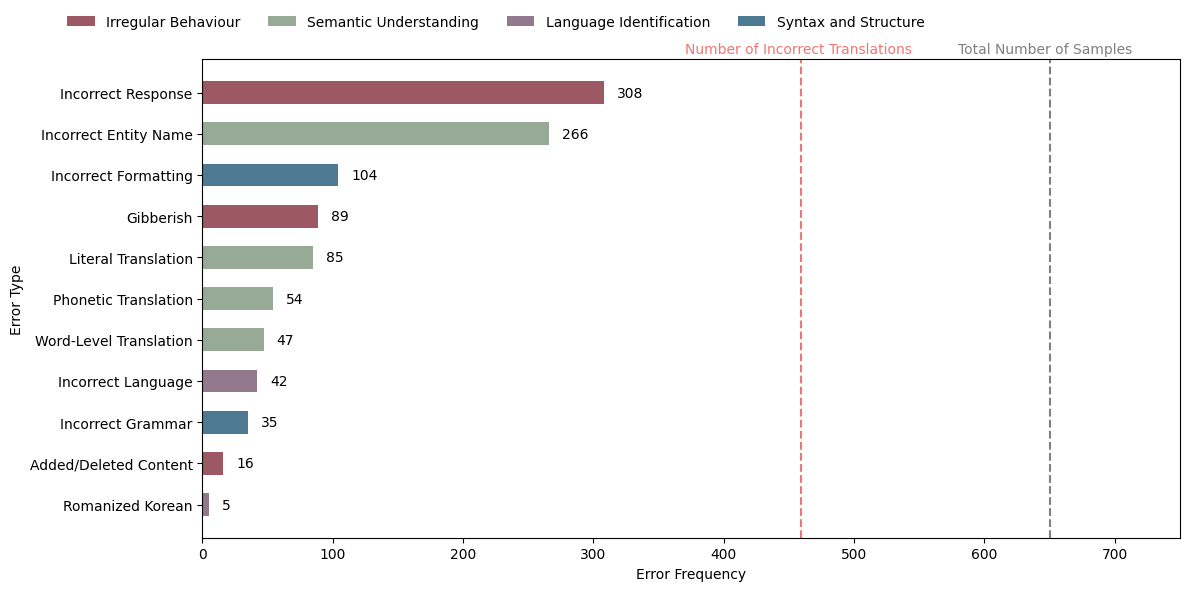}
    \caption{Frequency of errors per error taxonomy label. }
    \label{fig:error_frequency}
\end{figure*}
\subsection{Human Evaluation}
\label{humaneval}
For human evaluation, annotators were responsible for annotating 50 different English-Korean question pairs for each of the 13 models. They were tasked with evaluating (1) if the translation was correct, (2) which string in the English and Korean texts was mistranslated, and (3) a rationale for the mistranslation (Figure~\ref{fig:task_design}).
To complete the task, we recruited two annotators with native fluency in English and Korean. Considering the importance of being able to understand and identify cultural and language-specific nuances in both English and Korean, annotators were required to have lived in South Korea and the USA for a minimum of 5 years each, and underwent a comprehensive interview to qualify for the task. Each annotator was compensated \$150 for the completion of the entire task.

\section{Results and Discussions}
The automatic evaluation results presented in Table~\ref{tab:automaticeval} summarize the BLEU, COMET, and M-ETA scores across all 13 models. For BLEU, o1 demonstrates superior performance, while o1-mini excels in COMET metrics and Gemini 1.5 Pro achieves the highest M-ETA score. These scores are closely followed by Grok-2 and GPT-4o across all metrics. Generally, most LLMs outperform traditional multilingual translation models such as MBART-50 and NLLB-200, with notable exceptions being DeepSeek R1, Llama 3, and both Claude 3.5 variants (Haiku and Sonnet).

To complement our automatic evaluation, we conducted human assessments to gain a more nuanced understanding of translation quality across models. Our analysis reveals that 459 out of 650 evaluated samples contain translation errors, with Grok 2 exhibiting the lowest error rate. Among these mistranslations, 266 cases involve incorrectly translated entities, which annotators identified by highlighting discrepancies between the English source and Korean target texts.

We further constructed a comprehensive error ontology adapted and expanded from \cite{Popovic2018} with annotator-provided explanations, illustrated in Table~\ref{fig:error_frequency}. The predominant error categories are ``Incorrect Response'' (308 pairs) and ``Incorrect Entity Name'' (266 pairs). ``Incorrect Response'', which encompasses behaviors unrelated to translation (e.g., answering questions rather than translating content), is most common, despite using identical prompts across all models as shown in Figure ~\ref{fig:error_frequency} due to its simple task. ``Incorrect Entity Name'' confirm that translating entities in knowledge-intensive and entity-dense texts remains particularly challenging. The primary failure modes involve literal, phonetic, or word-for-word translations that fail to capture the semantic content of the source text, demonstrating limited cross-lingual comprehension of entities. 
The definitions for each error label can be found in Table~\ref{tab:error_translation_taxonomy}.

\section{Further Analysis}
We conduct a deeper, comprehensive analysis from our evaluation results to understand fine-grained insights in machine translation. In Section~\ref{popas}, we investigate whether entity popularity and typology influence translation quality. Section~\ref{autohum} explores the relationship between automatic metrics and human evaluations, providing insights into their complementary nature and potential discrepancies. 

\subsection{Impact of Entity Popularity and Type}
\label{popas}
\begin{figure*}[t]
        \centering
        \includegraphics[width=.8\textwidth]{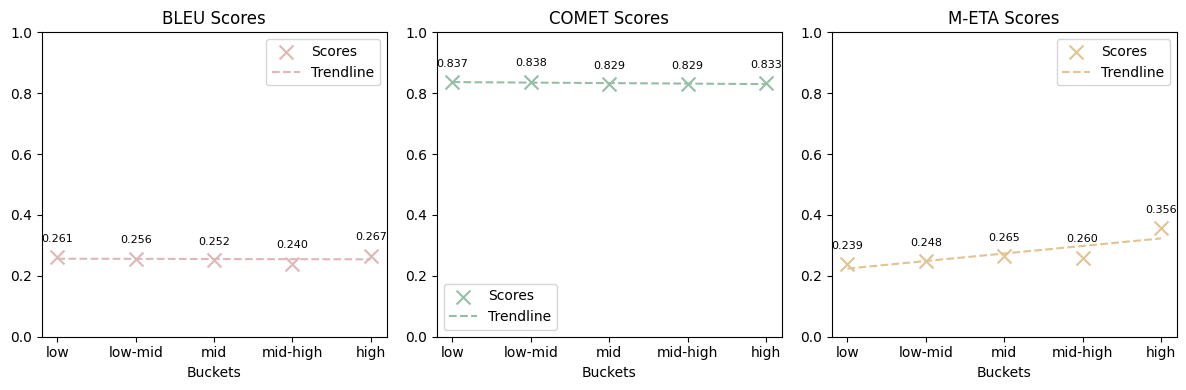}
    \caption{Average BLEU, COMET, M-ETA scores by popularity level.}
    \label{fig:average_graph}
\end{figure*}

We examine whether entity popularity influences machine translation quality, hypothesizing that frequently occurring entities in training data may yield better translations. Since the training corpora for these models are not publicly accessible, we use entity prominence as measured by Wikipedia page view statistics as a proxy for frequency in training data, operating under the assumption that widely-recognized entities are more likely to appear frequently in the data used to train these systems.
To test this, we categorized entities from our dataset into five popularity segments based on their 2024 page view counts on Wikipedia: Low, Low-Mid, Mid, Mid-High, and High. As illustrated in Table~\ref{tab:popularity_scores}, traditional metrics like BLEU and COMET remain relatively stable across these segments, with average scores of [0.26, 0.26, 0.25, 0.24, 0.27] and [0.84, 0.84, 0.83, 0.83, 0.83] respectively. However, M-ETA demonstrates a notable variation of 0.00224 across the popularity spectrum~(Figure~\ref{fig:average_graph}). Our findings suggest that while entity popularity impacts the translation quality of the entity itself, it does not significantly affect the translation of the surrounding sentence. Standard metrics such as BLEU and COMET fail to capture these nuances due to the relatively small token representation of entities within full sentences. This underscores the necessity for more fine-grained evaluation metrics that can assess culturally-nuanced and language-specific translation quality, rather than optimizing only for conventional translation metrics.

We further analyze performance by entity type (as categorized in Wikidata and standard named entity recognition (NER) types presented in \cite{tedeschi-etal-2021-wikineural-combined}) to investigate its influence on translation quality. Our results reveal performance disparities across different entity types in our dataset, with Plant and Natural place-related entities demonstrating higher performance across all 13 models. To distinguish between entity type effects and popularity level effects, we calculated the correlation between entity type performance and popularity scores~\ref{tab:entityByPopularity}. While the correlation patterns largely align with our previous findings, we observe several notable deviations.

Through qualitative analysis, we identify specific mechanisms by which entity types influence translation quality. For instance, entity type Book Series contains a higher concentration of names that could be simply literally, phonetically, or word-for-word translated, whereas entity type Plant and Natural place presents more challenges like requiring unique language-specific names. This suggests that translation difficulty is partially determined by the linguistic properties characteristic of specific entity categories, independent of their popularity or frequency in training data.

\subsection{Comparing Automatic and Human Evaluation}
\label{autohum}
By comparing BLEU, COMET, and M-ETA scores with human evaluation results, we investigate the correlation between automatic metrics and human judgment to identify aspects of translation quality that may not be captured by computational assessment. Our analysis reveals that BLEU and COMET demonstrate a moderately positive correlation with binary human evaluations of translation correctness (i.e., whether the translated text preserves the meaning and comprehensibility of the source). Across 650 annotated samples, we observe a point-biserial correlation coefficient of 0.41 with a p-value of 3.54e-28, indicating a moderately strong alignment between automatic metrics and human assessment. M-ETA correctly identifies 88.7\% samples containing entity translation errors according to human labels \cite{25bad119-8b17-35e3-84e1-8bd3b0f44e99}. 

\section{Conclusion and Future Work}
\label{conclusions}
In this paper, we evaluated 13 large language models and multilingual machine translation systems on their ability to handle culturally-nuanced and language-specific translation tasks across knowledge-intense and entity-dense questions from English to Korean. Our comprehensive assessment combined three automatic evaluation metrics with complementary human evaluations to thoroughly understand model performance. While our findings demonstrate that LLMs generally outperform traditional multilingual machine translation models, significant challenges remain, particularly regarding the appropriate transliteration versus transcreation of text. We hope this work encourages future research expanding beyond entity-dense and knowledge-intensive content to explore additional language pairs and text genres, ultimately informing targeted improvements in machine translation capabilities.

\section*{Limitations}
In this section, we discuss some of the limitations of our work and how future research may be able to address them.

\paragraph{Language and Dialect Coverage.}
This paper focuses on a detailed analysis of Korean (Koreanic), for its morphologically complex, typologically different translation from English (Indo-European). However, it lacks an investigation into other language families like Romance (e.g., Spanish, French), Semitic (e.g., Arabic), Altaic (e.g., Turkish) and more. Future work should focus on related in-depth analysis on other languages and dialects, to develop a robust and generalizable understanding of errors in the culturally-nuanced and language-specific machine translation of text.

\paragraph{Error Ontology Coverage.}
We acknowledge the limitations of the proposed ontology, as our evaluation was restricted to a controlled set of question templates with a predefined entity pool from only English to Korean. A broader analysis of diverse text genres, such as long-form documents or narrative content, would likely reveal additional error categories. Furthermore, our current error classification system does not account for the varying degrees to which translation errors impact semantic comprehension across source and target languages, which means the framework inadequately captures important dimensions of translation quality.

\section*{Acknowledgements}
We thank the organizers of SemEval 2025 Task 2: Entity-Aware Machine Translation (EA-MT) for creating a well-structured task. We also thank Simone Conia and Revanth Gangi Reddy for their valuable discussions and insightful feedback.

\newpage

\bibliography{acl_latex}

\appendix
\section{Appendix}
\label{sec:appendix}

\definecolor{bleuColor}{HTML}{E0B7B7}
\definecolor{cometColor}{HTML}{94BFA7}
\definecolor{metaColor}{HTML}{E2C391}

\input{tables/human_evaluation_table}

\input{tables/entity_popularity}

\input{tables/entity_scores_per_model}

\input{tables/error_translation_type}

\input{tables/popularity_table}

\begin{figure*}[t]
        \centering
        \includegraphics[width=.8\textwidth]{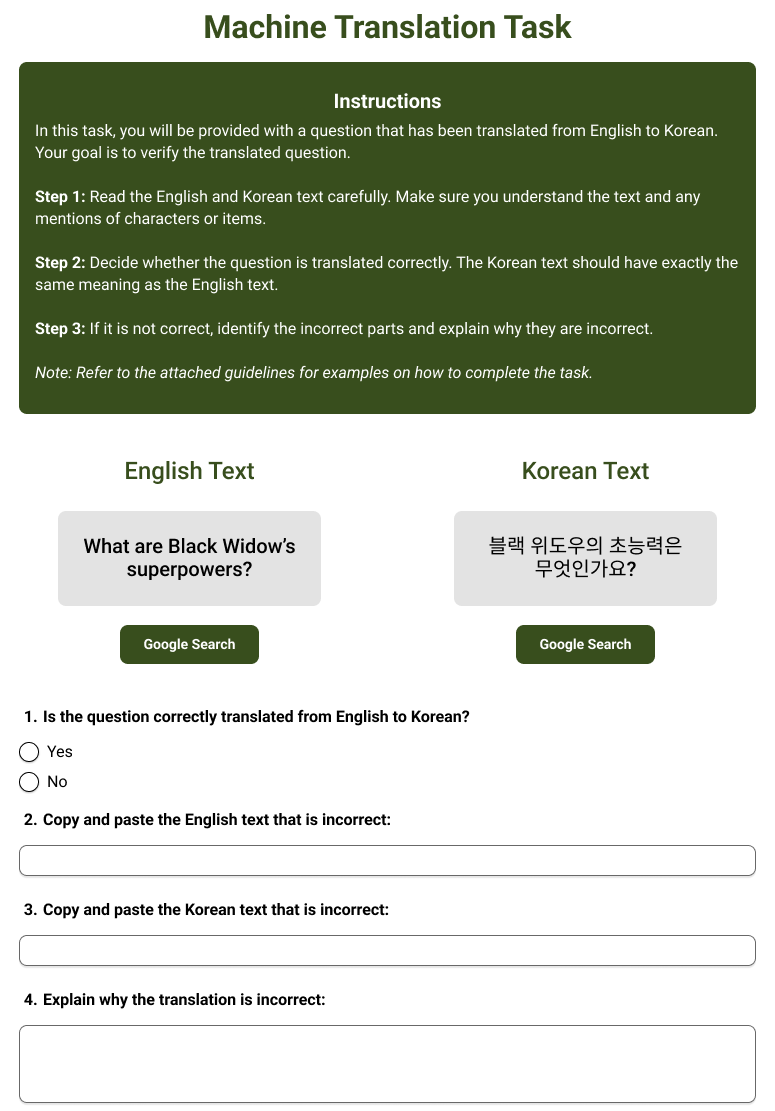}
    \caption{UI used for machine translation human annotation task.}
    \label{fig:task_design}
\end{figure*}

\end{document}

%% file: tables/automaticeval.tex
\definecolor{bleuColor}{HTML}{E0B7B7}
\definecolor{cometColor}{HTML}{94BFA7}
\definecolor{metaColor}{HTML}{E2C391}

\begin{table}[ht]
    \centering
    \resizebox{\columnwidth}{!}{%
    \begin{tabular}{llccc}
        \hline
        & & \multicolumn{3}{c}{\textbf{Metrics}} \\
        \textbf{Company} & \textbf{Models} & \it{BLEU} & \it{COMET} & \it{M-ETA} \\
        \hline
        \multirow{4}{*}{OpenAI} 
            & \it{o1} & \cellcolor{bleuColor} 0.3869 & 0.9196 & 0.3752 \\
            & \it{o1 Mini} & 0.3830 & \cellcolor{cometColor} 0.9202 & 0.3306 \\
            & \it{GPT-4o} & 0.3692 & 0.9087 & 0.3951 \\
            & \it{GPT-4o Mini} & 0.3545 & 0.9046 & 0.2914 \\
        \multirow{2}{*}{Anthropic} 
            & \it{Claude 3.5 Sonnet} & 0.1961 & 0.8384 & 0.3969 \\
            & \it{Claude 3.5 Haiku} & 0.1584 & 0.8056 & 0.2849 \\
        \multirow{2}{*}{Google} 
            & \it{Gemini 1.5 Pro} & 0.3810 & 0.9094 & \cellcolor{metaColor} 0.4833 \\
            & \it{Gemini 1.5 Flash} & 0.2965 & 0.9081 & 0.3316 \\
        xAI & \it{Grok 2} & 0.3808 & 0.9143 & 0.3514 \\
        DeepSeek & \it{DeepSeek R1} & 0.0066 & 0.4895 & 0.0026 \\
        \multirow{3}{*}{Meta} 
            & \it{Llama 3} & 0.0327 & 0.5529 & 0.0563 \\
            & \it{Mbart-50} & 0.1451 & 0.8702 & 0.0791 \\
            & \it{NLLB-200} & 0.2195 & 0.8899 & 0.1663 \\
        \hline
    \end{tabular}
    }
    \caption{Automatic evaluation results on BLEU, COMET, and M-ETA for English-Korean translation.}
    \textbf{Color key:} 
    \textcolor{black}{\fboxsep0.5pt\colorbox{bleuColor}{\rule{0pt}{6pt}\rule{6pt}{0pt}}} = Highest BLEU,
    \textcolor{black}{\fboxsep0.5pt\colorbox{cometColor}{\rule{0pt}{6pt}\rule{6pt}{0pt}}} = Highest COMET,
    \textcolor{black}{\fboxsep0.5pt\colorbox{metaColor}{\rule{0pt}{6pt}\rule{6pt}{0pt}}} = Highest M-ETA.
    \label{tab:automaticeval}
\end{table}

%% file: tables/human_evaluation_table.tex
\definecolor{bleuColor}{HTML}{E0B7B7}
\definecolor{cometColor}{HTML}{94BFA7}
\definecolor{metaColor}{HTML}{E2C391}
\definecolor{humanColor}{HTML}{5f8632}

\begin{table*}[ht]
    \centering
    \resizebox{0.8\textwidth}{!}{%
    \begin{tabular}{lcccc}
        \toprule
         & BLEU & COMET & M-ETA & Annotator Score \\
        \midrule
        \it{o1} & \cellcolor{bleuColor} 0.39 & 0.91 & 0.30 & 0.52 \\
        \it{o1 Mini} & \cellcolor{bleuColor} 0.39 & 0.91 & 0.38 & 0.50 \\
        \it{GPT-4o} & 0.38 & \cellcolor{cometColor}0.93 & 0.26 & 0.40 \\
        \it{GPT-4o Mini} & 0.34 & 0.92 & 0.34 & 0.30 \\
        \it{Claude 3.5 Sonnet} & 0.22 & 0.83 & 0.40 & 0.22 \\
        \it{Claude 3.5 Haiku} & 0.14 & 0.80 & 0.24 & 0.12 \\
        \it{Gemini 1.5 Pro} & \cellcolor{bleuColor} 0.39 & 0.90 & \cellcolor{metaColor}0.40 & 0.38 \\
        \it{Gemini 1.5 Flash} & 0.29 & 0.92 & 0.28 & 0.34 \\
        \it{Grok 2} & 0.32 & 0.92 & 0.36 & \cellcolor{humanColor} 0.58 \\
        \it{DeepkSeek R1} & 0.00 & 0.49 & 0.00 & 0.00 \\
        \it{Llama 3} & 0.04 & 0.58 & 0.04 & 0.06 \\
        \it{MBart05} & 0.17 & 0.87 & 0.10 & 0.22 \\
        \it{NLLB-200} & 0.22 & 0.88 & 0.22 & 0.14 \\
        \bottomrule
    \end{tabular}
    }
    \caption{Automatic evaluation results for BLEU, COMET and M-ETA and Human Evaluation Score.}
    \textbf{Color key:} 
    \textcolor{black}{\fboxsep0.5pt\colorbox{bleuColor}{\rule{0pt}{6pt}\rule{6pt}{0pt}}} = Highest BLEU,
    \textcolor{black}{\fboxsep0.5pt\colorbox{cometColor}{\rule{0pt}{6pt}\rule{6pt}{0pt}}} = Highest COMET,
    \textcolor{black}{\fboxsep0.5pt\colorbox{metaColor}{\rule{0pt}{6pt}\rule{6pt}{0pt}}} = Highest M-ETA.
    \textcolor{black}{\fboxsep0.5pt\colorbox{humanColor}{\rule{0pt}{6pt}\rule{6pt}{0pt}}} = Highest Annotator Score.
    \label{tab:results}
\end{table*}


%% file: tables/entity_popularity.tex
\definecolor{bleuColor}{HTML}{E0B7B7}
\definecolor{cometColor}{HTML}{94BFA7}
\definecolor{metaColor}{HTML}{E2C391}

\begin{table*}[ht]
    \centering
    \resizebox{0.8\textwidth}{!}{%
    \begin{tabular}{ccccc}
    \toprule
    Popularity Rank & Entity Type & BLEU & COMET & M-ETA \\
    \midrule
    1 & Plant & \cellcolor{bleuColor} 0.3448 & 0.8354 & \cellcolor{metaColor} 0.6573 \\
    2 & Book & 0.2245 & 0.8188 & 0.2721 \\
    3 & Person & 0.2666 & 0.8526 & 0.2704 \\
    4 & Artwork & 0.2096 & 0.8148 & 0.2497 \\
    5 & Food & 0.3317 & 0.8421 & 0.3646 \\
    6 & Movie & 0.1707 & 0.8151 & 0.1812 \\
    7 & Fictional entity & 0.3070 & 0.8337 & 0.3685 \\
    8 & Animal & 0.3026 & 0.8257 & 0.3846 \\
    9 & Landmark & 0.3026 & 0.8784 & 0.2552 \\
    10 & TV series & 0.2078 & 0.8077 & 0.1628 \\
    11 & Place of worship & 0.3141 & 0.8625 & 0.3070 \\
    12 & Natural place & 0.1638 & \cellcolor{cometColor} 0.9058 & 0.6410 \\
    13 & Musical work & 0.3297 & 0.8558 & 0.3735 \\
    14 & Book series & 0.2471 & 0.7992 & 0.0804 \\
    \bottomrule
    \end{tabular}
    }
    \caption{Entites ranked by popularity levels along with their average scores.}
    \textbf{Color key:} 
    \textcolor{black}{\fboxsep0.5pt\colorbox{bleuColor}{\rule{0pt}{6pt}\rule{6pt}{0pt}}} = Highest BLEU,
    \textcolor{black}{\fboxsep0.5pt\colorbox{cometColor}{\rule{0pt}{6pt}\rule{6pt}{0pt}}} = Highest COMET,
    \textcolor{black}{\fboxsep0.5pt\colorbox{metaColor}{\rule{0pt}{6pt}\rule{6pt}{0pt}}} = Highest M-ETA.
    \label{tab:entityByPopularity}
\end{table*}

%% file: tables/entity_scores_per_model.tex
\definecolor{bleuColor}{HTML}{E0B7B7}
\definecolor{cometColor}{HTML}{94BFA7}
\definecolor{metaColor}{HTML}{E2C391}

\begin{table*}[ht]
\centering
\resizebox{\textwidth}{!}{%
\begin{tabular}{cc|cccccccccccccc}
\toprule
& & \multicolumn{14}{c}{\textbf{Entity Types}} \\ 
    \textbf{Models} & \textbf{Metric} & Animal & Artwork & Book & Book Series & Fictional Entity & Food & Landmark & Movie & Musical Work & Natural place & Person & Place of Worship & Plant & TV Series \\
\midrule

\multirow{3}{*}{\it{o1}}
  & BLEU
    & 0.3712 & 0.3236 & 0.3334 & \cellcolor{bleuColor} 0.3964
    & 0.4659 & 0.4988 & 0.4284
    & 0.2953 & 0.4477 & 0.2108
    & \cellcolor{bleuColor} 0.4243 & \cellcolor{bleuColor} 0.5019 & \cellcolor{bleuColor} 0.7097 & \cellcolor{bleuColor} 0.3202 \\
  & Comet
    & 0.9393 & 0.9018 & 0.9029 & 0.8779
    & \cellcolor{cometColor} 0.9384 & \cellcolor{cometColor} 0.9438 & 0.9537
    & 0.8965 & \cellcolor{cometColor} 0.9388 & 0.9720
    & \cellcolor{cometColor} 0.9426 & \cellcolor{cometColor} 0.9488 & \cellcolor{cometColor} 0.9582 & 0.8862 \\
  & M-ETA
    & 0.5000 & 0.3602 & 0.3634 & 0.1940
    & \cellcolor{metaColor} 0.5703 & 0.5092 & 0.3413
    & 0.2683 & 0.3912 & 1.0000
    & 0.4118 & 0.4448 & 0.8182 & 0.2396 \\
\midrule

\multirow{3}{*}{\it{o1 Mini}}
  & BLEU
    & 0.3698 & 0.3204 & 0.3504 & 0.3827
    & \cellcolor{bleuColor} 0.5013 & 0.5301 & \cellcolor{bleuColor} 0.4352
    & 0.2143 & 0.4869 & 0.2927
    & 0.4034 & 0.4563 & 0.5268 & 0.3186 \\
  & Comet
    & 0.9223 & \cellcolor{cometColor} 0.9063 & \cellcolor{cometColor} 0.9115 & 0.8770
    & 0.9359 & 0.9430 & \cellcolor{cometColor} 0.9540
    & \cellcolor{cometColor} 0.8991 & 0.9385 & \cellcolor{cometColor} 0.9768
    & 0.9410 & 0.9435 & 0.9324 & 0.8838 \\
  & M-ETA
    & 0.0000 & 0.2762 & 0.3225 & 0.0448
    & 0.4934 & 0.4479 & 0.2857
    & 0.1847 & 0.5705 & 1.0000
    & 0.3069 & 0.3785 & 0.7273 & 0.1506 \\
\midrule

\multirow{3}{*}{\it{GPT-4o}}
  & BLEU
    & \cellcolor{bleuColor} 0.5247 & 0.2926 & 0.3204 & 0.3116
    & 0.4544 & \cellcolor{bleuColor} 0.5301 & 0.4332
    & 0.2423 & 0.4644 & 0.2717
    & 0.4211 & 0.4491 & 0.6030 & 0.2835 \\
  & Comet
    & 0.9046 & 0.8938 & 0.8975 & 0.8629
    & 0.9122 & 0.9418 & 0.9495
    & 0.8858 & 0.9217 & 0.9684
    & 0.9276 & 0.9363 & 0.9420 & 0.8801 \\
  & M-ETA
    & 0.6667 & 0.3782 & 0.4091 & 0.1493
    & 0.5093 & 0.4847 & 0.3651
    & 0.2840 & 0.5034 & 0.6667
    & 0.3890 & 0.4038 & 0.9091 & 0.2927 \\
\midrule

\multirow{3}{*}{\it{GPT-4o Mini}}
  & BLEU
    & 0.4254 & 0.2755 & 0.3049 & 0.3569
    & 0.4230 & 0.4827 & 0.4144
    & 0.2517 & 0.4520 & 0.3137
    & 0.3789 & 0.4261 & 0.5220 & 0.2968 \\
  & Comet
    & 0.8876 & 0.8782 & 0.8786 & 0.8737
    & 0.9090 & 0.9374 & 0.9487
    & 0.8943 & 0.9202 & 0.9720
    & 0.9213 & 0.9346 & 0.9496 & 0.8845 \\
  & M-ETA
    & 0.1667 & 0.2736 & 0.2960 & 0.0597
    & 0.4164 & 0.4387 & 0.3016
    & 0.1829 & 0.3735 & 0.6667
    & 0.2692 & 0.3596 & 0.9091 & 0.1664 \\
\midrule

\multirow{3}{*}{\it{Claude-3.5 Sonnet}}
  & BLEU
    & 0.1324 & 0.1507 & 0.1685 & 0.1644
    & 0.2000 & 0.2349 & 0.2578
    & 0.1257 & 0.3372 & 0.0000
    & 0.1681 & 0.2243 & 0.2190 & 0.1674 \\
  & Comet
    & 0.7358 & 0.8193 & 0.8329 & 0.8183
    & 0.7991 & 0.8123 & 0.8787
    & 0.8466 & 0.8939 & 0.8295
    & 0.8340 & 0.8463 & 0.7539 & 0.8269 \\
  & M-ETA
    & 0.8333 & 0.3448 & 0.3682 & 0.0597
    & 0.4934 & \cellcolor{metaColor} 0.5429 & 0.3333
    & 0.2526 & 0.6142 & 0.6667
    & 0.3688 & \cellcolor{metaColor} 0.4479 & \cellcolor{metaColor} 1.0000 & 0.2439 \\
\midrule

\multirow{3}{*}{\it{Claude-3.5 Haiku}}
  & BLEU
    & 0.0407 & 0.1320 & 0.1377 & 0.1921
    & 0.1568 & 0.2241 & 0.2089
    & 0.1129 & 0.2528 & 0.0000
    & 0.1255 & 0.1492 & 0.1348 & 0.1331 \\
  & Comet
    & 0.6727 & 0.7875 & 0.7851 & 0.7962
    & 0.7663 & 0.7913 & 0.8724
    & 0.7834 & 0.8557 & 0.8263
    & 0.8035 & 0.8346 & 0.7320 & 0.7981 \\
  & M-ETA
    & 0.8333 & 0.2787 & 0.2960 & 0.1045
    & 0.3979 & 0.4755 & 0.2937
    & 0.2073 & 0.2599 & \cellcolor{metaColor} 1.0000
    & 0.2948 & 0.3880 & 0.8182 & 0.1535 \\
\midrule
\multirow{3}{*}{\it{Gemini-1.5-Pro}}
  & BLEU
    & 0.4971 & \cellcolor{bleuColor} 0.3514 & \cellcolor{bleuColor} 0.3689 & 0.2932
    & 0.4189 & 0.4678 & 0.4132
    & \cellcolor{bleuColor} 0.2972 & 0.4914 & 0.2381
    & 0.3786 & 0.4714 & 0.4760 & 0.2703 \\
  & Comet
    & \cellcolor{cometColor} 0.9508 & 0.8927 & 0.8920 & 0.8578
    & 0.9166 & 0.9371 & 0.9501
    & 0.8812 & 0.9371 & 0.9667
    & 0.9291 & 0.9412 & 0.9522 & 0.8739 \\
  & M-ETA
    & 0.5000 & \cellcolor{metaColor} 0.4871 & \cellcolor{metaColor} 0.5174 & \cellcolor{metaColor} 0.1940
    & 0.5570 & 0.5245 & \cellcolor{metaColor} 0.3889
    & \cellcolor{metaColor} 0.3345 & \cellcolor{metaColor} 0.7291 & 0.0000
    & \cellcolor{metaColor} 0.4616 & 0.4227 & 0.8182 & \cellcolor{metaColor} 0.3286 \\
\midrule
\multirow{3}{*}{\it{Gemini-1.5-Flash}}
  & BLEU
    & 0.4989 & 0.2424 & 0.2656 & 0.2561
    & 0.3868 & 0.4001 & 0.3482
    & 0.2291 & 0.3242 & 0.0000
    & 0.3308 & 0.3846 & 0.5315 & 0.2332 \\
  & Comet
    & 0.9499 & 0.8876 & 0.8932 & 0.8548
    & 0.9207 & 0.9292 & 0.9459
    & 0.8907 & 0.9248 & 0.9634
    & 0.9342 & 0.9393 & 0.9532 & 0.8721 \\
  & M-ETA
    & \cellcolor{metaColor} 0.8333 & 0.3045 & 0.3586 & 0.0896
    & 0.4695 & 0.4417 & 0.3175
    & 0.2439 & 0.3748 & 0.3333
    & 0.3513 & 0.4006 & 0.9091 & 0.2023 \\

\midrule

\multirow{3}{*}{\it{Grok 2}}
  & BLEU
    & 0.3493 & 0.2808 & 0.3042 & 0.3665
    & 0.4617 & 0.4973 & 0.4121
    & 0.2571 & 0.5490 & 0.3137
    & 0.3912 & 0.4749 & 0.4891 & 0.3171 \\
  & Comet
    & 0.9053 & 0.8923 & 0.8980 & \cellcolor{cometColor} 0.8876
    & 0.9250 & 0.9387 & 0.9499
    & 0.8902 & 0.9356 & 0.9747
    & 0.9358 & 0.9443 & 0.9516 & \cellcolor{cometColor} 0.8885 \\
  & M-ETA
    & 0.1667 & 0.3113 & 0.3430 & 0.1343
    & 0.5013 & 0.4417 & 0.2460
    & 0.2213 & \cellcolor{bleuColor} 0.5636 & 1.0000
    & 0.3163 & 0.3849 & 0.9091 & 0.2023 \\

\midrule

\multirow{3}{*}{\it{DeepSeek-R1-7B}}
  & BLEU
    & 0.0315 & 0.0076 & 0.0076 & 0.0085
    & 0.0034 & 0.0060 & 0.0137
    & 0.0049 & 0.0071 & 0.0000
    & 0.0044 & 0.0045 & 0.0000 & 0.0090 \\
  & Comet
    & 0.5619 & 0.4891 & 0.4971 & 0.4952
    & 0.4752 & 0.4664 & 0.4914
    & 0.5036 & 0.4911 & 0.5595
    & 0.4858 & 0.4714 & 0.4533 & 0.5038 \\
  & M-ETA
    & 0.0000 & 0.0026 & 0.0036 & 0.0000
    & 0.0053 & 0.0031 & 0.0079
    & 0.0000 & 0.0000 & 0.0000
    & 0.0013 & 0.0000 & 0.0000 & 0.0072 \\
\midrule

\multirow{3}{*}{\it{Llama 3}}
  & BLEU
    & 0.0128 & 0.0154 & 0.0193 & 0.0407
    & 0.0566 & 0.0601 & 0.0642
    & 0.0130 & 0.0310 & 0.0000
    & 0.0450 & 0.0570 & 0.0000 & 0.0238 \\
  & Comet
    & 0.5322 & 0.4972 & 0.4997 & 0.5121
    & 0.5633 & 0.5500 & 0.6507
    & 0.5124 & 0.5784 & 0.8461
    & 0.6225 & 0.6564 & 0.4895 & 0.5111 \\
  & M-ETA
    & 0.0000 & 0.0223 & 0.0409 & 0.0000
    & 0.1114 & 0.1258 & 0.0952
    & 0.0261 & 0.0328 & 0.6667
    & 0.0983 & 0.0946 & 0.0000 & 0.0316 \\
\midrule

\multirow{3}{*}{\it{mBART-Large-50}}
  & BLEU
    & 0.2799 & 0.1540 & 0.1573 & 0.1808
    & 0.1977 & 0.1655 & 0.1982
    & 0.0716 & 0.1282 & 0.1291
    & 0.1684 & 0.1949 & 0.0529 & 0.1125 \\
  & Comet
    & 0.8659 & 0.8647 & 0.8692 & 0.8330
    & 0.8814 & 0.8848 & 0.9301
    & 0.8330 & 0.8897 & 0.9638
    & 0.8878 & 0.9007 & 0.8821 & 0.8341 \\
  & M-ETA
    & 0.0000 & 0.0823 & 0.0927 & 0.0000
    & 0.1088 & 0.1227 & 0.1746
    & 0.0488 & 0.0643 & 1.0000
    & 0.0888 & 0.0726 & 0.2727 & 0.0488 \\
\midrule

\multirow{3}{*}{\it{NLLB-200}}
  & BLEU
    & 0.3999 & 0.1781 & 0.1805 & 0.2620
    & 0.2638 & 0.2141 & 0.3061
    & 0.1044 & 0.3141 & \cellcolor{bleuColor} 0.3591
    & 0.2258 & 0.2895 & 0.2180 & 0.2159 \\
  & Comet
    & 0.9060 & 0.8813 & 0.8870 & 0.8429
    & 0.8948 & 0.8719 & 0.9447
    & 0.8800 & 0.8998 & 0.9566
    & 0.9183 & 0.9146 & 0.9107 & 0.8576 \\
  & M-ETA
    & 0.5000 & 0.1244 & 0.1252 & 0.0149
    & 0.1565 & 0.1810 & 0.1667
    & 0.1010 & 0.3776 & 0.3333
    & 0.1575 & 0.1924 & 0.4545 & 0.0488 \\
\bottomrule
\end{tabular}
}
\caption{BLEU, COMET, and M-ETA scores for each model and entity type.}
    \textbf{Color key:} 
    \textcolor{black}{\fboxsep0.5pt\colorbox{bleuColor}{\rule{0pt}{6pt}\rule{6pt}{0pt}}} = Highest BLEU,
    \textcolor{black}{\fboxsep0.5pt\colorbox{cometColor}{\rule{0pt}{6pt}\rule{6pt}{0pt}}} = Highest COMET,
    \textcolor{black}{\fboxsep0.5pt\colorbox{metaColor}{\rule{0pt}{6pt}\rule{6pt}{0pt}}} = Highest M-ETA.
\label{tab:entity_scores_per_model}
\end{table*}

%% file: tables/error_translation_type.tex
\begin{table*}[ht]
    \centering
    \begin{tabular}{ll}
        \toprule
        \textbf{Type of Error} & \textbf{Definition} \\
        \midrule
        Literal Translation & Translation follows the meaning of the source language. \\
        Phonetic Translation & Translation follows how it sounds in the source language. \\
        Word-Level Translation & Translation is done word-for-word from the source language. \\
        Incorrect Entity Name & Used a different or less appropriate entity name. \\
        Incorrect Grammar & Grammar mistakes in the target language. \\
        Incorrect Language & Translated into the wrong language. \\
        Incorrect Formatting & Formatting is wrong, but the translation itself is correct. \\
        Added/Deleted Content & Extra parts added or parts missing compared to the source. \\
        Incorrect Response & Output that doesn't match the source text meaning. \\
        Partial Translation & Only part of the source text is translated. \\
        Romanized Korean & Latin alphabet used instead of proper Korean script. \\
        Gibberish & Output makes no sense at all. \\
        \bottomrule
    \end{tabular}
    \caption{Types of translation errors and their definitions.}
    \label{tab:error_translation_taxonomy}
\end{table*}

%% file: tables/popularity_table.tex
\definecolor{bleuColor}{HTML}{E0B7B7}
\definecolor{cometColor}{HTML}{94BFA7}
\definecolor{metaColor}{HTML}{E2C391}

\begin{table*}[ht]
    \centering
    \begin{minipage}{\textwidth}
        \centering
        \resizebox{\textwidth}{!}{%
        \begin{tabular}{cc|ccccc}
    \toprule
    & & \multicolumn{5}{c}{\textbf{Popularity Level}} \\
    \textbf{Models} & \textbf{Metric} & \textbf{Low} & \textbf{Low-mid} & \textbf{Mid} & \textbf{Mid-high} & \textbf{High} \\
    & & \it{142 - 12943} & \it{12944 - 29440} & \it{29441 - 62685} & \it{62686 - 157350} & \it{157351 - 6974823}\\
    \midrule
    \multirow{3}{*}{\it{o1}} & BLEU  & 0.3789 & 0.3764 & 0.3760 & \cellcolor{bleuColor} 0.3696 & \cellcolor{bleuColor} 0.4377 \\
    & COMET & 0.9167 & 0.9173 & \cellcolor{cometColor} 0.9149 & 0.9184 & 0.9297 \\
    & M-ETA & 0.3171 & 0.3259 & 0.3415 & 0.3579 & 0.5321 \\
    \midrule
    \multirow{3}{*}{\it{o1 Mini}} & BLEU  & 0.3883 & \cellcolor{bleuColor} 0.3791 & 0.3723 & 0.3621 & 0.4180 \\
    & COMET & \cellcolor{cometColor} 0.9179 & \cellcolor{cometColor} 0.9200 & 0.9145 & \cellcolor{cometColor} 0.9186 & \cellcolor{cometColor} 0.9300 \\
    & M-ETA & 0.3028 & 0.3005 & 0.3129 & 0.3013 & 0.4455 \\
    \midrule
    \multirow{3}{*}{\it{GPT-4o}} & BLEU  & 0.3713 & 0.3689 & 0.3676 & 0.3496 & 0.3910 \\
    & COMET & 0.9082 & 0.9122 & 0.9036 & 0.9094 & 0.9109 \\
    & M-ETA & 0.3618 & 0.3492 & 0.4006 & 0.3589 & 0.5066 \\
    \midrule
    \multirow{3}{*}{\it{GPT-4o Mini}} & BLEU  & 0.3720 & 0.3615 & 0.3373 & 0.3449 & 0.3607 \\
    & COMET & 0.9078 & 0.9082 & 0.8960 & 0.9058 & 0.9059 \\
    & M-ETA & 0.2530 & 0.2680 & 0.2681 & 0.2841 & 0.3894 \\
    \midrule
    \multirow{3}{*}{\it{Claude 3.5 Sonnet}} & BLEU  & 0.2088 & 0.2114 & 0.1996 & 0.1672 & 0.1977 \\
    & COMET & 0.8490 & 0.8530 & 0.8374 & 0.8264 & 0.8285 \\
    & M-ETA & 0.3567 & 0.3777 & 0.3812 & 0.3761 & 0.4995 \\
    \midrule
    \multirow{3}{*}{\it{Claude 3.5 Haiku}} & BLEU  & 0.1881 & 0.1794 & 0.1557 & 0.1332 & 0.1393 \\
    & COMET & 0.8269 & 0.8219 & 0.8032 & 0.7918 & 0.7829 \\
    & M-ETA & 0.2266 & 0.2538 & 0.2803 & 0.2781 & 0.3914 \\
    \midrule
    \multirow{3}{*}{\it{Gemini 1.5 Pro}} & BLEU  & 0.3707 & 0.3776 & 0.3743 & 0.3691 & 0.4239 \\
    & COMET & 0.9108 & 0.9127 & 0.8995 & 0.9066 & 0.9183 \\
    & M-ETA & \cellcolor{metaColor} 0.4360 & \cellcolor{metaColor} 0.4589 & \cellcolor{metaColor} 0.4638 & \cellcolor{metaColor} 0.4611 & \cellcolor{metaColor} 0.5994 \\
    \midrule
    \multirow{3}{*}{\it{Gemini 1.5 Flash}} & BLEU  & 0.2864 & 0.2827 & 0.2948 & 0.2923 & 0.3305 \\
    & COMET & 0.9079 & 0.9117 & 0.9035 & 0.9058 & 0.9120 \\
    & M-ETA & 0.2571 & 0.2964 & 0.3425 & 0.3195 & 0.4444 \\
    \midrule
    \multirow{3}{*}{\it{Grok 2}} & BLEU  & \cellcolor{bleuColor} 0.4061 & 0.3779 & \cellcolor{bleuColor} 0.3919 & 0.3498 & 0.3869 \\
    & COMET & 0.9130 & 0.9157 & 0.9118 & 0.9135 & 0.9173 \\
    & M-ETA & 0.3018 & 0.3147 & 0.3405 & 0.3488 & 0.4648 \\
    \midrule
    \multirow{3}{*}{\it{DeepSeek R1}} & BLEU  & 0.0091 & 0.0087 & 0.0051 & 0.0052 & 0.0041 \\
    & COMET & 0.4955 & 0.4924 & 0.4852 & 0.4837 & 0.4892 \\
    & M-ETA & 0.0041 & 0.0020 & 0.0020 & 0.0020 & 0.0020 \\
    \midrule
    \multirow{3}{*}{\it{Llama 3}} & BLEU  & 0.0369 & 0.0281 & 0.0376 & 0.0315 & 0.0286 \\
    & COMET & 0.5652 & 0.5634 & 0.5513 & 0.5373 & 0.5466 \\
    & M-ETA & 0.0539 & 0.0315 & 0.0550 & 0.0586 & 0.0765 \\
    \midrule
    \multirow{3}{*}{\it{MBart-50}} & BLEU  & 0.1446 & 0.1411 & 0.1464 & 0.1453 & 0.1442 \\
    & COMET & 0.8700 & 0.8742 & 0.8675 & 0.8660 & 0.8735 \\
    & M-ETA & 0.0640 & 0.0558 & 0.0744 & 0.0809 & 0.1172 \\
    \midrule
    \multirow{3}{*}{\it{NLLB-200}} & BLEU  & 0.2317 & 0.2399 & 0.2236 & 0.2006 & 0.2042 \\
    & COMET & 0.8915 & 0.8933 & 0.8871 & 0.8912 & 0.8884 \\
    & M-ETA & 0.1667 & 0.1848 & 0.1784 & 0.1547 & 0.1600 \\
    \bottomrule
    \end{tabular}        
        }
        \caption{BLEU, COMET and M-ETA scores grouped by popularity levels for each model.}
        \textbf{Color key:} 
        \textcolor{black}{\fboxsep0.5pt\colorbox{bleuColor}{\rule{0pt}{6pt}\rule{6pt}{0pt}}} = Highest BLEU,
        \textcolor{black}{\fboxsep0.5pt\colorbox{cometColor}{\rule{0pt}{6pt}\rule{6pt}{0pt}}} = Highest COMET,
        \textcolor{black}{\fboxsep0.5pt\colorbox{metaColor}{\rule{0pt}{6pt}\rule{6pt}{0pt}}} = Highest M-ETA.
        \label{tab:popularity_scores}
    \end{minipage}
\end{table*}